\documentclass[11pt]{article}

\usepackage[final]{acl}

\usepackage{times}
\usepackage{latexsym}

\usepackage[T1]{fontenc}

\usepackage[utf8]{inputenc}

\usepackage{microtype}

\usepackage{inconsolata}

\usepackage{graphicx}
\usepackage{booktabs}
\usepackage{tabularx}
\usepackage{graphicx} 
\usepackage{multirow}
\usepackage{listings}
\usepackage[section]{placeins} 
\usepackage{hyperref}

\title{ReFRAME or Remain: Unsupervised Lexical Semantic Change Detection with Frame Semantics}

\author{
  \textbf{Bach Phan-Tat}\textsuperscript{1},
  \textbf{Kris Heylen}\textsuperscript{1,2},
  \textbf{Dirk Geeraerts}\textsuperscript{1},
  \textbf{Stefano De Pascale}\textsuperscript{1,3},
  \textbf{Dirk Speelman}\textsuperscript{1}
  \\[1ex]
  \textsuperscript{1}Department of Linguistics, KU Leuven\\
  \textsuperscript{2}Instituut voor de Nederlandse Taal\\
  \textsuperscript{3}Vrije Universiteit Brussel\\[1ex]
  \small \textbf{Correspondence:} \href{mailto:ttbach.phan@kuleuven.be}{ttbach.phan@kuleuven.be}
}

\begin{document}
\maketitle
\begin{abstract}
The majority of contemporary computational methods for lexical semantic change (LSC) detection are based on neural embedding distributional representations. Although these models perform well on LSC benchmarks, their results are often difficult to interpret. We explore an alternative approach that relies solely on frame semantics. We show that this method is effective for detecting semantic change and can even outperform many distributional semantic models. Finally, we present a detailed quantitative and qualitative analysis of its predictions, demonstrating that they are both plausible and highly interpretable.
\end{abstract}

\section{Introduction}

Lexical semantic change detection is well-established in Natural Language Processing (NLP), with different shared tasks for different languages, amongst which the most popular one is SemEval 2020 Task 1 \cite{schlechtweg_semeval-2020_2020}. Most modern systems are based on either static word embeddings like word2vec \cite{mikolov_efficient_2013} or ELMo \cite{peters_deep_2018}, or contextualised ones such as BERT \cite{devlin_bert_2019} or S-BERT \cite{reimers_sentence-bert_2019}, the latter currently achieving the state-of-the-art (SOTA) results \cite{cassotti_xl-lexeme_2023}. Despite their architectural differences, these methods are grounded in the distributional hypothesis \cite{firth_synopsis_1957}, and can capture some aspects of lexical meaning \citep[e.g.,][]{iacobacci_embeddings_2016, lenci_comparative_2022, pilehvar_wic_2019, yenicelik_how_2020}. Using them for modelling semantic change would therefore be the logical next step.

\begin{figure}[htbp]
    \centering
    \includegraphics[width=1\linewidth]{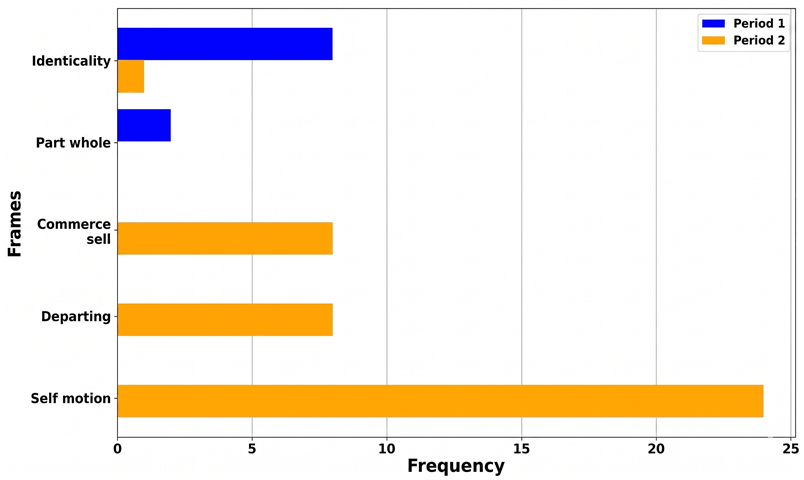}
    \caption{Changes in the frequencies of 6 frames that the English noun \textit{plane} participated in between the 2 sub\-corpora for English of the SemEval 2020 shared task 1 \cite{schlechtweg_semeval-2020_2020}. \textit{Plane} is annotated as semantically changed in the dataset}
    \label{fig:plane_frame}
\end{figure}

However, co-occurrence is only one perspective of lexical semantics. There are other perspectives/frameworks such as Behavioural Profiles \cite{evans_behavioral_2009}, Frame Semantics \cite{fillmore_frame_1982}, Conceptual Metaphor \cite{lakoff_metaphors_1981} (see \citet{geeraerts_theories_2010} for a comprehensive overview of lexical semantics), and some of them have also been used in diachronic lexical semantics and semantic change research \citep[e.g.,][]{geeraerts_prototype_1983, jansegers_towards_2020, law_diachronic_2019, pettersson-traba_analyzing_2016, sovran_polysemy_2004, sullivan_metaphoric_2007, sweetser_semantic_1990, vais_diachronic_2022}.

Exploiting Frame Semantics for detecting lexical semantic change is the focus of this paper. Consider the English noun \textit{plane}: in the 20\textsuperscript{th} century, after the birth of the modern airplane in 1903, the \textsc{aircraft} meaning emerged and became more dominant while the \textsc{dimensional} meaning became less dominant. This is reflected in the shift in the frame distributions of \textit{plane}, where \textsc{aircraft}-related frames such as \textit{commerce\_sell}, \textit{departing}, \textit{self\_motion} increase in frequency and \textsc{dimensional}-related frames such as \textit{identicality}, \textit{part\_whole} decrease in frequency, as shown in Figure \ref{fig:plane_frame}. 

Our main hypothesis is that \textbf{semantic change is reflected in changes in the frames in which a word participates, either as a frame-evoking element or as a frame element}. This view is compatible with distributional approaches: large-scale distributional models may also capture aspects of frame structure implicitly. The contrast lies instead in the level of representation. Rather than relying on implicit information in the embedding space, we use explicitly annotated frame-semantic information as a simple and transparent signal for lexical semantic change. Our aim is therefore not to establish a new state of the art, but to draw attention to interpretable methods that make use of explicit linguistic information.

We investigate this question experimentally\footnote{All the codes for parsing, preprocessing and experimenting are included in \url{https://github.com/phantatbach/STARSEM26}.} using the English dataset\footnote{We could only test this method on the English dataset due to the lack of available and reliable frame semantic resources for other languages. This is justified in section \ref{sec:data_and_tasks}.} from SemEval 2020 Task 1. Our main findings are the following:
\begin{itemize}
    \item Tracing the changes in the distribution of frames outperforms many distributional semantic models, including both static embeddings (count-based and prediction-based) and contextualised embeddings.
    \item Because the method is interpretable by design, we can decompose change scores into contributions of individual frames and inspect supporting contexts, enabling targeted qualitative analyses to understand how a word has changed (as in the plane example above) in contrast to the opaqueness of dense embedding models\footnote{While we acknowledge the growing field of explainable AI and its diverse range of techniques (e.g., Integrated Gradients \cite{sundararajan_axiomatic_2017}, LIME \cite{ribeiro_why_2016} or SHAP \cite{lundberg_unified_2017}, see surveys in \citet{arrieta_explainable_2019, madsen_post-hoc_2022}), these methods are often post-hoc, add methodological and computational overhead, and may not yield explanations that are straightforward to inspect linguistically. This makes our approach particularly convenient for linguistic and humanities research, where direct, feature-level inspection is desired.}.
    \item Our approach is one of the attempts to bridge NLP methods with linguistic theories of lexical semantics and semantic change \citep[e.g.,][]{dalpanagioti_corpus_2019, geeraerts_diachronic_1997, geeraerts_theories_2010, geeraerts_lexical_2023, hanks_lexical_2013}.
\end{itemize}

\section{Related work}
A long-standing tradition in diachronic semantics builds on the Distributional Hypothesis \cite{firth_synopsis_1957}. With the rise of corpus linguistics and concordancing, researchers began tracing a target word’s collocates and recurring patterns across corpora and time slices, often using association measures to identify characteristic collocations \cite{church_word_1990, sinclair_corpus_1991}. This collocational approach has been applied in many case studies \cite[e.g.,][]{alba-salas_life_2007, pettersson-traba_measuring_2021}, methods \cite{tat-etal-2026-transparent, garcia_method_2019, mcenery_usage_2019}, and tools \cite{anthony_antconc_2014, brezina_lancsbox_2020, kilgarriff_sketch_2014, jurish_diacollo_2015}. With advances in computational modelling, distributional information was represented as numerical vectors, from count-based matrices \cite{bullinaria_extracting_2007, gulordava_distributional_2011, schutze_automatic_1998} to dense neural embeddings \cite{baroni_dont_2014, bojanowski_enriching_2017, levy_dependency-based_2014, mikolov_efficient_2013, pennington_glove_2014} and contextualised token-level models \cite{devlin_bert_2019}. In lexical semantic change research, standard workflows typically split corpora into time slices, learn and align representations, and measure change through distances or clustering \cite{tahmasebi_survey_2021, tahmasebi_computational_2023}. As shared tasks and standardised datasets became central, compact vector representations and single change scores became especially attractive, and now dominate much of the field, including current SOTA approaches \cite[e.g.,][]{cassotti_xl-lexeme_2023, geeraerts_lexical_2023, giulianelli_analysing_2020, kim_temporal_2014, kulkarni_statistically_2015, kutuzov_diachronic_2018, montariol_scalable_2021, tahmasebi_computational_2023, tahmasebi_survey_2021}.

Other theory-driven frameworks have also been employed in Diachronic Semantics \cite[e.g.,][]{geeraerts_prototype_1983, jansegers_towards_2020, law_diachronic_2019, pettersson-traba_analyzing_2016, sovran_polysemy_2004, sullivan_metaphoric_2007, sweetser_semantic_1990, vais_diachronic_2022}. However, these approaches often yield descriptive outputs, making it difficult to compare systematically across languages and datasets. Recently, efforts have been made to benchmark these theory-driven approaches with standard shared tasks (e.g., \citeauthor{kutuzov_grammatical_2021}, \citeyear{kutuzov_grammatical_2021}; \citeauthor{ryzhova_detection_2021}, \citeyear{ryzhova_detection_2021} on grammatical profiles; \citeauthor{tang_can_2023}, \citeyear{tang_can_2023} on sense distributions). Their strong performances on shared-task benchmarks (sometimes even outperform many embedding-based methods) inspired us to test the effectiveness of using Frame Semantics information in capturing word meaning change.

Frame Semantics was introduced by \citet{fillmore_frame_1982} as a theory of meaning in which words are understood relative to structured background knowledge called frames. A frame is a conceptual scene (e.g., \textit{Commercial\_transaction, Motion}) that specifies the key participants, relations, and typical inferences involved. The meaning of a lexical unit is not just a standalone definition, but a particular way of evoking a frame and profiling parts of it. For example, \textit{buy, sell, pay}, and \textit{cost} all relate to the same \textit{commercial} frame, but each highlights different roles and perspectives. Frames come with frame elements that represent the core components of the scene (e.g., \textit{buyer, seller, goods, money}). The theory is operationalised in resources such as FrameNet \cite{baker_framenet_2014}, which catalogs frames, frame elements, and the lexical units that evoke them, along with annotated examples. Because of its structured representation of ‘who did what to whom’ at the level of conceptual situations, Frame Semantic representations have supported applications in semantic parsing, information extraction, and linguistic analysis \cite[e.g.,][]{law_diachronic_2019, li_joint_2019, marzinotto_semantic_2018, robin_identifying_2023, vais_diachronic_2022}.

Some practical weaknesses of Frame Semantics are:
\begin{itemize}
    \item The substantial time and manual work needed for building corpora.
    \item The lack of coverage for different languages as frame inventories are built from particular languages and cultural/institutional settings, although efforts have been made for cross-lingual applicability \cite{boas_framenet_2025}.
    \item The lack of coverage for historical and out-of-domain data as many senses, lexical items, constructions are missing from the inventory \cite{hartmann_out--domain_2017, palmer_evaluating_2010}.
\end{itemize} These may explain why it has mainly been used in specific linguistics case studies \cite{blank_why_1999, blank_words_2003, georgakopoulos_frame-based_2018, koch_cognitive_1999, koch_cognitive_2008, koch_frame_1999, law_diachronic_2019, vais_diachronic_2022} and not in unsupervised computational approaches to LSC detection. As far as we know, our work is the first one to employ Frame Semantics for this task.

\section{Data and tasks}
\label{sec:data_and_tasks}

Following SemEval 2020 Task 1 on Unsupervised Lexical Semantic Change Detection \cite{schlechtweg_semeval-2020_2020}, we formulate the problem as either binary classification or graded ranking, corresponding to subtasks 1 and 2. In subtask 1, given a set of target words, the system predicts whether each word gained or lost at least one sense between two time periods. In subtask 2, the system ranks target words by their degrees of semantic change. Annotating semantic change is challenging because it requires assessing meaning differences across many usage instances from each period. A widely used protocol is DURel \cite{schlechtweg_diachronic_2018}, where annotators compare pairs of sentences containing a target word and judge whether the instances are used in the same sense or in different senses. These pairwise judgments are then aggregated into a final change score, either binary or continuous. This procedure was used to create the SemEval 2020 Task 1 dataset, the most widely used one to compared different LSC detection systems. It covers four languages: English, German, Latin, and Swedish, with manual annotations for both subtasks, and provides a diachronic corpus split into two time periods for each language.  Using this benchmark enables direct comparison of our approach with SOTA systems under a shared evaluation setup.

\begin{table*}[!t]
    \centering
    \begin{tabular}{l c l l c}
    \toprule
    \textbf{Period} & \textbf{Tokens} & \textbf{Types} & \textbf{Corpus} & \textbf{Target words} \\ 
    \midrule
    C1 (1810 -- 1860) & 6.5M & Raw, Lemma & CCOHA & \multirow{2}{*}{37} \\
    C2 (1960 -- 2010) & 6.7M & Raw, Lemma & CCOHA & \\ 
    \bottomrule
    \end{tabular}
    \caption{Descriptions of the English dataset in SemEval 2020 Task 1.}
    \label{tab:eng_semeval}
\end{table*}

However, our approach requires high-quality frame-semantic parses, which are not currently available for all languages. While English benefits from extensive FrameNet resources and several mature automatic frame-semantic parsers, we were unable to find comparably reliable off-the-shelf parsers for the other 3 languages. For English, we used the pretrained Frame Semantic Transformer model \cite{chanin_open-source_2023}, which is trained on FrameNet 1.7 and is reported to achieve SOTA performance. It is released as a Python library, making it convenient for using and reproducing the results. For German, although there are several FrameNet-like resources, they were all relatively dated \cite{boas_bilingual_2002, burchardt_8_2009, burchardt_salsa_2006, erk_shalmaneser_2006}. The most recent resource and parser we have found is PFN-DE \cite{bick_framenet_2022}. However, it is only accessible through a hosted interface that only accepts text files up to 10,000 characters, and processing larger volumes requires contacting the provider for a paid processing agreement \cite{noauthor_visl_2026}. This makes it impractical to parse 140 million token historical German corpora \cite{schlechtweg_semeval-2020_2020} at scale compared to a locally runnable, openly distributed pipeline, and raises concern about transparency and reproducibility. Moreover, it has only been evaluated on modern Twitter data, making its reliability on historical German uncertain. For Swedish, Svenskt frasnät 2.0 (SweFN) Dataset \cite{sprakbanken_text_svenskt_2024} provides a FrameNet-style resource. The latest attempt to build a pretrained Swedish frame parser with SweFN used the same architecture as the English ones mentioned above, but with substantially lower performance \cite{dannells_frame-semantic_2025, dannells_transformer-based_2024}. In a small sanity check with 30 random, diverse and basic sentences, we also observed 13 questionable frame predictions, such as \textit{Cutting} for \textit{Jag såg en film igår}‘I saw a movie yesterday’ (see the results in table \ref{tab:test_swe_en} in the Appendix). The low F1 scores, coupled with its failure in such simple examples, raise concern about the reliability of the model. For Latin, we did not find any public frame-semantic resources or pretrained parsers.

Because of the lack of available and quality frame-semantic parsers for German, Swedish and Latin, we restrict the experiment to English where strong pretrained parsers exist. We compensate for the lack of multilingual quantitative experiments with a detailed qualitative analysis. A quick description of the English data is presented in Table \ref{tab:eng_semeval}. Note that although we evaluate only English, frame semantics is in principle language-agnostic \cite{fillmore_frame_1982} so the approach should generalise once suitable frame resources and parsers exist.

\section{Method}

\subsection{FrameNet parsing}
Because we only need sentences containing the target lemma, we first split both the lemmatised and raw data into lemma-specific subcorpora. Since the lemmatised and raw files contain the same number of sentences in the same order, we do this by scanning the lemmatised file to select sentences that include the target lemma, then retrieving the corresponding sentences from the raw (token) file by index. We then frame-parsed them using the Frame transformer model \cite{chanin_open-source_2023}. The parser is available in two variants, \textsc{base} and \textsc{small}, with similar reported performances. It has also been noted that the models can occasionally identify a frame but return empty outputs. We therefore use \textsc{base} by default and fall back to \textsc{small} when it fails. If both models fail, we skip the sentence. Across 49,416 sentences, we triggered the fallback for 2,413 lemmatised sentences (4.9\%) and 1,052 raw sentences (2.1\%). Amongst these, both models failed (and the sentences were skipped) for only 805 lemmatised sentences (1.6\%) and 336 raw sentences (0.6\%).

One technical limitation of the parser is that it returns the frame elements as raw strings without locations. This makes it impossible to locate the target lemma in the parsed token data (one target lemma may appear in different surface forms). We therefore use lemmatised corpora rather than raw corpora. To assess the impact of doing so, we compared (within each target word and time period) the frame distributions derived from the lemmatised and raw corpora using Jensen–Shannon Divergence \cite{menendez_jensen-shannon_1997}. The resulting JSD values (in the table \ref{tab:token_vs_lemma} in the Appendix) are uniformly small (all < 0.15, most < 0.10), indicating that using lemmatised corpora introduces only minor changes to the induced frame distributions.

\subsection{Basic procedure}
To obtain the frame distributions, we collected all annotated frame instances in which the target lemma occurs either as the frame-evoking element or within the lexical span of one of the frame elements. 
\begin{itemize}
\item Each frame instance was counted at most once. Even if the target lemma appeared both as the frame trigger and inside a frame element, or appeared in multiple frame-element spans of the same frame instance, that frame instance was still counted only once. This de-duplication was applied only within individual frame instances. 
\item If the target lemma participated in multiple frame instances in the same sentence, all such frame instances were retained and counted as separate occurrences.
\end{itemize}

We then experiment with 2 types of frame distributions\footnote{We intentionally excluded the frame – trigger only distribution. This is because some lemmas only appear in frame elements and not evoke any frame, making using frame – triggers only yield sparse or even empty distributions for some targets, causing JSD to be unstable or even undefined.}: frame–elements only (FE) and frame–triggers + frame–elements (FTFE). To quantify the distributional changes across periods, we use Jensen–Shannon divergence (JSD) \cite{menendez_jensen-shannon_1997}. JSD is a symmetric, smoothed variant of Kullback–Leibler Divergence \cite{kullback_information_1951}, and with log base 2 it is bounded between 0 and 1, making scores easy to interpret and comparable across targets. In addition, JSD can be decomposed into per-item contributions, allowing fine-grained attribution of which items (in this case, frames) drive the observed change. Prior work on grammatical profiles \cite{kutuzov_grammatical_2021, ryzhova_detection_2021} quantified changes with cosine distance between time-specific vectors, but cosine compares vector geometry (angles) rather than directly comparing probability distributions.

\section{Results}
\label{sec:results}
We evaluate our method on both subtasks of SemEval 2020 Task 1. As described in Section \ref{sec:data_and_tasks}, subtask 1 is a binary classification task evaluated by accuracy, while subtask 2 is a ranking task evaluated by Spearman’s rank correlation between the JSDs and the GOLD SCORES. Our primary focus is Subtask 2. For Subtask 1, we classify a word as \textsc{changed} if its JSD >= 0.5, else \textsc{unchanged}.

\subsection{Subtask 2}
The results for subtask 2 are reported in table \ref{tab:subtask_2}. Using frame-elements only, we were able to achieve a spearman correlation of 0.249. This is already much higher than the Count Baseline (0.022) and surpassing many static and contextualised embedding systems. Using the combined frame-triggers and frame-elements boosts the score to 0.306, making our system amongst the top 10 systems for English. This is very impressive given the simplicity of the approach.

\begin{table}[!t]
\centering
\begin{tabular}{l c l}
\toprule
\textbf{Team} & \textbf{EN score} & \textbf{Type} \\
\midrule
XL-Lexeme & 0.757 & token \\
NLPCR & 0.436 & token \\
UG\_Student\_Intern & 0.422 & type \\
cs2020 & 0.375 & type \\
UWB & 0.367 & type \\
Discovery\_Team & 0.361 & ens.\ \\
Jiaxin \& Jinan & 0.325 & type \\
UCD & 0.307 & graph \\
\textbf{FrameNet-FTFE} & \textbf{0.306} & -- \\
IMS & 0.301 & type \\
Life-Language & 0.299 & type \\
Syntactic Dependency & 0.277 & -- \\
Entity & 0.250 & type \\
\textbf{FrameNet-FE} & \textbf{0.249} & -- \\
RPI-Trust & 0.228 & type \\
Grammatical Profiling & 0.218 & -- \\
Random & 0.211 & type \\
Skurt & 0.209 & token \\
RIJP & 0.157 & type \\
UiO-UvA & 0.136 & token \\
UoB & 0.105 & topic \\
cbk & 0.059 & token \\
NLP@IDSIA & 0.028 & token \\
\textbf{Count Bas.} & 0.022 & -- \\
JCT & 0.014 & type \\
TUE & -0.155 & token \\
DCC & -0.217 & type \\
\textbf{Freq. Bas.} & -0.217 & -- \\
\textbf{Maj. Bas.} & -- & -- \\
\bottomrule
\end{tabular}
\caption{Performances of our models in comparison with other systems in graded change detection (SemEval 2020 Subtask 2), Spearman rank correlation coefficients. FTFE = Frame-Trigger-Frame-Element, FE = Frame Element.}
\label{tab:subtask_2}
\end{table}

\subsection{Subtask 1}
The results for subtask 1 are presented in table \ref{tab:subtask_1}. As the experiment set-up is fully unsupervised, different teams come up with different strategies to classify the changing status of the target lemmas. We based our binary classification score assignment on the 0.5 cutoff point of JSD. The threshold of 0.5 was chosen empirically, based on our prior experience using JSD on linguistic data. It acts as a high-pass filter for semantic salience: it highlights substantial structural shifts while suppressing minor fluctuations that often arise from sparse distributions. We avoid any supervised finetuning for selecting the optimal JSD threshold to preserve the fully unsupervised spirit of the task and comparability with other unsupervised approaches. Both of our systems perform equally well, with an accuracy of 0.622, on par with different word embedding systems.

\section{Qualitative analysis}
A key strength of our system is its interpretability. JSD decomposition reveals which distributional shifts drive the change, and users can inspect the corresponding contexts to locate the underlying semantic change. In this section, we qualitatively analyse the predictions of our better system (FTFE) to investigate what has been captured.

\begin{table}[!t]
\centering
\begin{tabular}{l c l}
\toprule
\textbf{Team} & \textbf{EN score} & \textbf{System type} \\
\midrule
NLPCR & 0.730 & token \\
Life-Language & 0.703 & type \\
Entity & 0.676 & type \\
Jiaxin \& Jinan & 0.649 & type \\
RPI-Trust & 0.649 & type \\
DCC & 0.649 & type \\
JCT & 0.649 & type \\
Syntactic Dependency & 0.649 & -- \\
\textbf{FrameNet-FTFE} & \textbf{0.622} & -- \\
\textbf{FrameNet-FE} & \textbf{0.622} & -- \\
UWB & 0.622 & type \\
Grammatical Profiling & 0.622 & -- \\
NLP@IDSIA & 0.622 & token \\
UCD & 0.622 & graph \\
\textbf{Count Bas.} & 0.595 & -- \\
cs2020 & 0.595 & token \\
UG\_Student\_Intern & 0.568 & type \\
Skurt & 0.568 & token \\
Discovery\_Team & 0.568 & ens.\ \\
TUE & 0.568 & token \\
\textbf{Maj. Bas.} & 0.568 & -- \\
cbk & 0.568 & token \\
UoB & 0.568 & topic \\
IMS & 0.541 & type \\
UiO-UvA & 0.541 & token \\
RIJP & 0.541 & type \\
Random & 0.486 & type \\
\textbf{Freq. Bas.} & 0.432 & -- \\
\bottomrule
\end{tabular}
\caption{Performances of our models in comparison with other systems in binary change detection (SemEval 2020 Subtask 1), accuracy. Note that in this paper we mostly focus on ranking (Subtask 2). FTFE = Frame-Trigger-Frame-Element, FE = Frame-Element.}
\label{tab:subtask_1}
\end{table}

We follow the same strategy as in our previous study \cite{tat-etal-2026-transparent}, where we divided the target lemmas into 5 groups: True Positive (TP), True Negative (TN), False Positive (FP), False Negative (FN), MID. We define TP as lemmas that fall in the top 33\% of both the Subtask 2 Gold Scores and our system’s predicted rankings, and TN as those in the bottom 33\% of both. FP are lemmas in the bottom 33\% of the gold ranking but the top 33\% of our ranking, and FN are lemmas in the top 33\% of the gold ranking but the bottom 33\% of our ranking. All remaining lemmas are assigned to a MID group. Our qualitative analysis focuses on TP, TN, FP, and FN, as these categories most clearly expose the method’s strengths and limitations. Due to space constraints, we present one illustrative (most extreme) example per group.  The lemmas of each group are listed in Table \ref{tab:qualitative_grouping} in the Appendix.

\begin{figure}
    \centering
    \includegraphics[width=1\linewidth]{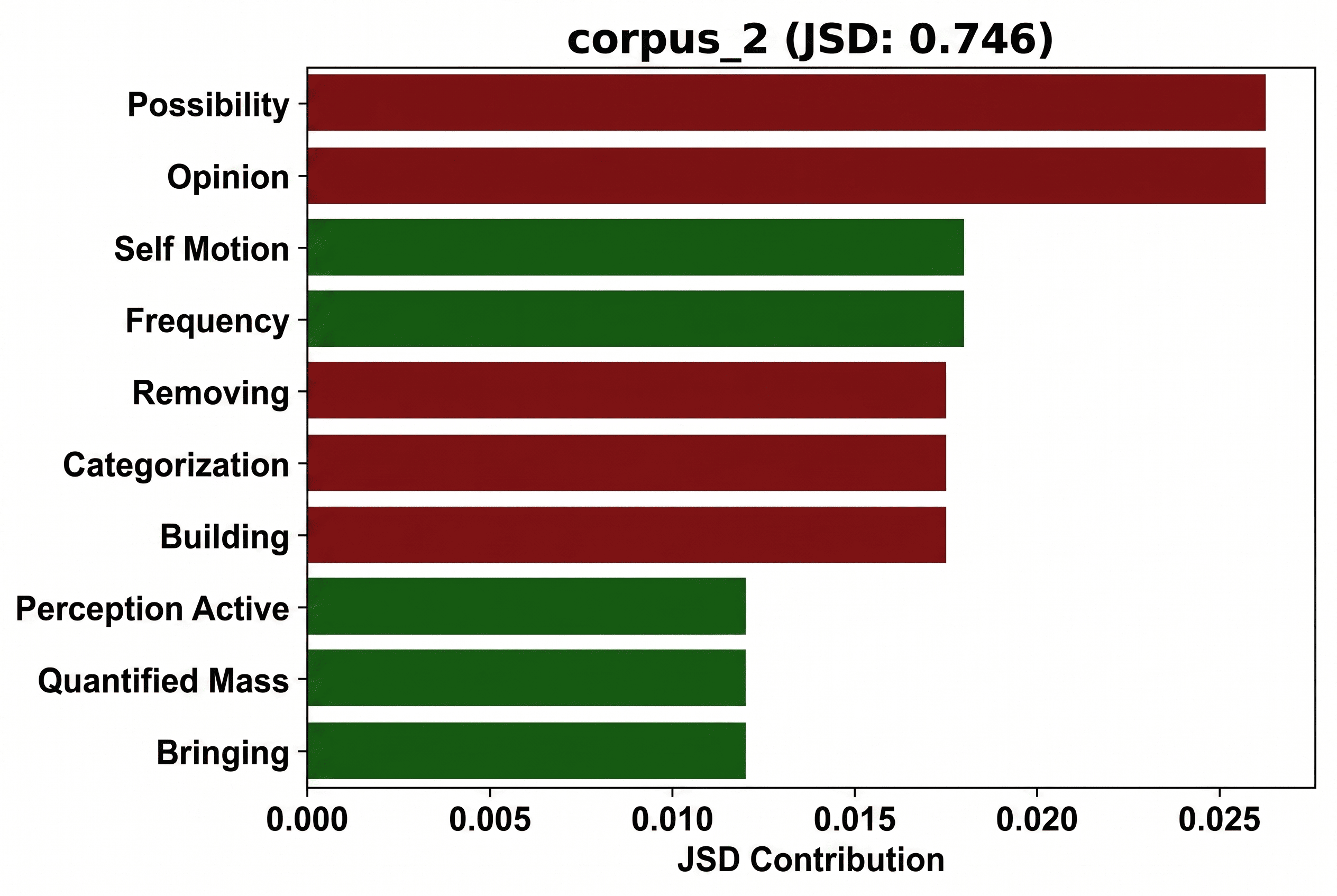}
    \caption{The JSD contributions of different frames of the English noun \textit{prop}. Green bars indicate an increase in relative frequency; red bars indicate a decrease.}
    \label{fig:prop_jsd}
\end{figure}

\paragraph{True positives} Upon qualitative analysis, semantic changes do indeed link to changes in frame distributions. The English noun \textit{prop} develops a new \textsc{theatre object} meaning whereas in the earlier period the \textsc{support} meaning (either literal or figurative) was dominant. In the first period, \textit{prop} were mostly linked to such frames as \textit{Possibility} (e.g., the possibility of finding support) or \textit{Opinion} (e.g, thinking about the pillar/bread-winner of the family, or opposing the election of the prop/council of the city) or Building. In the second period, those frames decrease and there was an increase in \textit{Self motion} frames, where \textit{prop} acts as the mover (e.g., \textit{… props fly into scene}) or looking for the \textit{prop} is the purpose of the mover (e.g., \textit{he clambered … looking for perfect props}) or \textit{Quantified\_mass} where \textit{prop} was quantified by \textit{few, many} or \textit{Bringing} where \textit{prop} is the object being brought. This is illustrated in figure \ref{fig:prop_jsd}.

\begin{figure}
    \centering
    \includegraphics[width=1\linewidth]{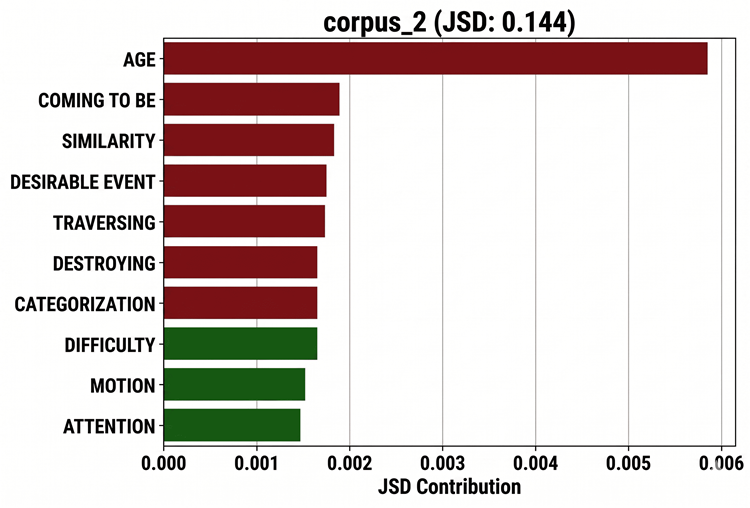}
    \caption{The JSD contributions of different frames of the English noun \textit{tree}. Green bars indicate an increase in relative frequency; red bars indicate a decrease.}
    \label{fig:tree_jsd}
\end{figure}

\paragraph{True negatives} The JSD scores of TNs are really low, suggesting that there are minor changes in the frames they appear in. The major shift of the English noun \textit{tree} (see Figure \ref{fig:tree_jsd}) is the fact the phrase \textit{old tree} appears less (C1: 76 times, C2: 9 times). Other shifts are minor (hence the small overall JSD) and simply contextual (e.g., the decrease in \textit{trees} being destroyed), not related to any real semantic change.

\begin{figure}
    \centering
    \includegraphics[width=1\linewidth]{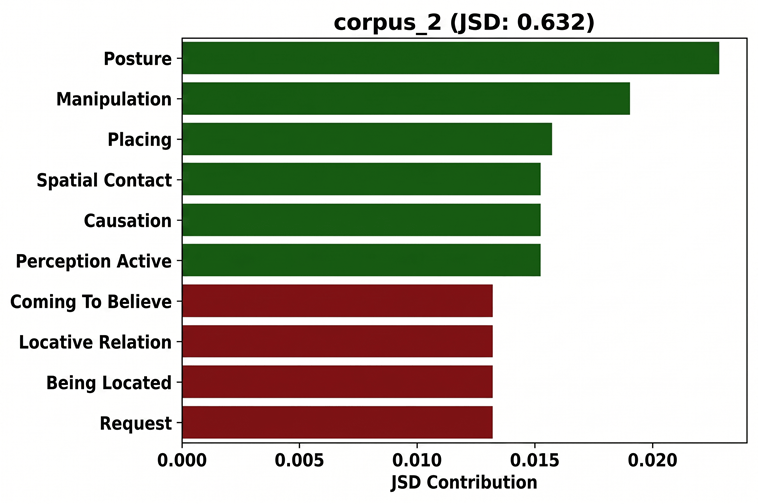}
    \caption{The JSD contributions of different frames of the English noun \textit{quilt}. Green bars indicate an increase in relative frequency; red bars indicate a decrease.}
    \label{fig:quilt_jsd}
\end{figure}

\paragraph{False positives} We only found one FP (\textit{quilt\_nn}). Our method detects large shifts in its frame distributions, yet it is ranked low in the gold Subtask 2 ranking. This is because of the large shift in the contexts (which do not correspond to any semantic change) in which \textit{quilt} appears in. In the second period, \textit{quilt} was the object of many activities, such as \textit{lean into the quilt. wring the quilt, lay the folded quilt}, etc. This results in a large increase of such frames as \textit{Posture, Manipulation, Placing, Spatial contact, Causation}, which drives the overall JSD up (see figure \ref{fig:quilt_jsd}).

\begin{figure}
    \centering
    \includegraphics[width=1\linewidth]{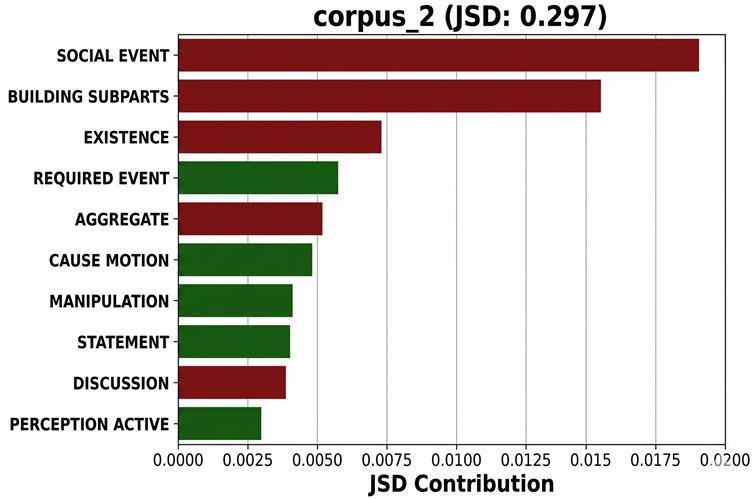}
    \caption{The JSD contributions of different frames of the English noun \textit{ball}. Green bars indicate an increase in relative frequency; red bars indicate a decrease.}
    \label{fig:ball_jsd}
\end{figure}

\paragraph{False negative} FNs are target words whose frame JSD scores underestimate the degree of change relative to the gold ranking. For example, our system does capture the shift for the English noun \textit{ball} from a usage dominated by \textsc{social event} to one dominated by \textsc{spherical object}, reflected in decreases in frames such as \textit{Social event} and \textit{Building subpart} (e.g., \textit{ball room}) and increases in frames such as \textit{Cause motion} and \textit{Manipulation} (see figure \ref{fig:ball_jsd}. We hypothesise that, for FN cases, these distributional shifts are just simply not large enough to place the target high in the ranked list. Nevertheless, the method remains useful for qualitative analysis, since it highlights the relevant frame shifts and supports interpretation of semantic change through the lens of Frame Semantics.

There are two common problems we observe during the qualitative analysis. First, some frames are highly general and carry little specific semantic content. For instance, in the \textit{ball} case, the rise of the \textit{Statement} frame largely reflects the frequent occurrences of \textit{say} in contexts that mention \textit{ball} (whether referring to the social event or the object), rather than a substantive meaning shift. Second, using lemmatised corpora can occasionally disrupt grammatical agreement or sentence boundaries, which may confuse the parser and lead to incorrect frame assignments (this effect appears minor overall, as shown by the low JSD between the raw and lemmatised corpora’s frame inventories).

\section{Conclusion}
This paper, along with other linguistically motivated case studies on SemEval 2020 Task 1 \cite[e.g.,][]{tat-etal-2026-transparent, kutuzov_grammatical_2021, ryzhova_detection_2021, tang_can_2023} demonstrates that explicit linguistic evidence is sufficient to build an effective lexical semantic change detection system. Our method achieves competitive performance on English and can outperform many distributional baselines on both subtasks. Although it does not match the current SOTA systems which are based on large pretrained language models and can encode a broader range of linguistic and contextual signals, maximising benchmark scores has never been our primary goal. Rather, we aim to highlight and thus bring more attention to theory-driven, interpretable approaches to lexical semantic change detection.

To further test and refine this approach, future work should:
\begin{itemize}
    \item Develop Frame inventories and more reliable Frame parsers for other languages
    \item Work around the problem of not being able to locate the locations of frame elements.
\end{itemize}

More broadly, we view this approach and some previous ones \cite[e.g.,][]{kutuzov_grammatical_2021, tat-etal-2026-transparent, ryzhova_detection_2021} as instances of a general framework for semantic change analysis: identify interpretable, theory-driven dimensions of a target’s distribution (here, frames) and quantify change within each dimension. This framework could (and should) be used as a complementary tool to interpret and validate the changes detected by neural networks.

\section*{Limitations}
We highlight two primary limitations and practical considerations of our proposed method.

First, frame inventory and frame parsers availability are a hard constraint, making this method inapplicable to languages lacking robust frame parsers. Even for English, the frame parser is still not perfect. It cannot output the token offsets for frame elements, meaning we must use the lemmatised version of the corpora, which can (minorly) affect the performance of the approach.

Second, our qualitative analysis, while supporting the interpretability claim, simultaneously reveals one inherent limitation of distributional frame tracking. The method can overestimate changes when large shifts in a target’s contexts do not correspond to genuine semantic change. Conversely, it can underestimate change when genuine semantic innovations are not sufficiently reflected in frame distributional shifts.

\section*{Ethical statement}
The data and all codes used in this study are publicly available. No human data was used, and no human participant was involved in the experiment.

\section*{Acknowledgments}
All ideas and core content were developed by the authors then later polished and optimised using ChatGPT and Gemini. We reviewed all AI-generated content thoroughly and take full responsibility for its accuracy.

This project has received funding from the European Union’s Horizon Europe programme for research and innovation under MSCA Doctoral Networks 2022, Grant Agreement No. 101120349 and Grant Agreement No. 101119511.

We would like to thank the anonymous reviewers for their detailed and informative comments and
suggestions.

\bibliography{custom}

\clearpage 
\onecolumn 
\appendix
\section*{Appendix}
\label{sec:appendix}

\section{Testing the Swedish model on 30 sentences}
\begin{table}[htbp]
    \centering    
    \resizebox{\textwidth}{!}{
        \begin{tabularx}{1.2\textwidth}{l l X X} 
            \toprule
            \textbf{Swedish} & \textbf{English translation} & \textbf{Predicted verbal frames - SWE} & \textbf{Predicted verbal frames - ENG} \\ 
            \midrule
            Jag såg en film igår & I saw a movie yesterday & \textbf{Cutting} & Perception\_experience \\
            Vi äger ett hus & We own a house & \textbf{Cause\_to\_fragment} & Possession \\
            Vi lånar pengar & We borrow money & Borrowing & Getting \\
            Mamma skickar ett brev & Mom sends a letter & Passing & Sending \\
            Barnet dricker mjölk & The child drinks milk & Ingestion & Ingestion \\
            Hon säger sanningen & She tells the truth & \textbf{Deciding} & Telling \\
            Jag vet svaret & I know the answer & \textbf{Health\_status} & Awareness \\
            Bilen stannar här & The car stops here & Halt & Halt \\
            Han öppnar fönstret & He opens the window & \textbf{Closure} & Closure \\
            Vi går till parken & We go to the park & Motion & Motion \\
            Tåget anländer nu & The train arrives now & Arriving & Arriving \\
            Boken ligger på hyllan & The book is on the shelf & \textbf{Posture} & No verbal frame. Only Text triggered by \textit{book} \\
            Jag tänker på dig & I think of you & Cogitation & Awareness \\
            Han älskar sin hund & He loves his dog & Experiencer\_focus & Experiencer\_focus \\
            Hon förstår frågan & She understands the question & \textbf{Frugality} & Awareness \\
            Vi glömmer namnet & We forget the name & Memory & Remembering\_information \\
            De pratar om vädret & They talk about the weather & Chatting & Chatting \\
            Läraren svarar eleven & The teacher answers the student & \textbf{Rising\_to\_a\_challenge} & Education\_teaching \\
            Pappa lovar en present & Dad promises a present & Commitment & \textbf{Cause\_to\_perceive} \\
            Hon ringer sin vän & She calls her friend & \textbf{Cause\_to\_make\_noise} & Referring\_by\_name \\
            Isen smälter snabbt & Ice melts quickly & Change\_of\_phase & Change\_of\_phase \\
            Solen skiner idag & The sun shines today & Light\_movement & Light\_movement \\
            Han slår igen dörren & He slams the door & \textbf{Fall\_for} & Cause\_impact \\
            Han skriver ett mejl & He writes an email & Text\_creation & Text\_creation \\
            Mötet börjar klockan nio & The meeting starts at nine o'clock & Process\_start & Process\_start \\
            Jag slår på TV:n & I turn on the TV & \textbf{Make\_noise} & Change\_operational\_state \\
            Det regnar ute & It is raining outside & Precipitation & Precipitation \\
            Barnet sover djupt & The child sleeps deeply & Sleep & Sleep \\
            Hon kommer på svaret & She comes up with the answer & \textbf{Satisfying} & Coming\_up\_with \\
            Hon ger upp & She gives up & \textbf{Excreting} & EMPTY FRAME \\ 
            \bottomrule
        \end{tabularx}
    }
    \caption{Testing the Swedish models on 30 random and basic sentences, amongst which 13 frames were questionable (in bold). Compare with the results from the English BASE model..}
    \label{tab:test_swe_en}
\end{table}

\section{JSD between frame distributions of raw and lemmatised English corpora}
\label{sec:appendix_token_vs_lemma}

\begin{table}[htbp]
\centering
\begin{tabular}{l c c}
\toprule
\textbf{Target lemma} & \textbf{JSD period 1 (token vs lemma)} & \textbf{JSD period 2 (token vs lemma)} \\
\midrule
attack\_nn & 0.053278 & 0.036771 \\
bag\_nn & 0.093656 & 0.049945 \\
ball\_nn & 0.077025 & 0.076959 \\
bit\_nn & 0.073138 & 0.048374 \\
chairman\_nn & 0.086464 & 0.042575 \\
circle\_vb & 0.118216 & 0.127109 \\
contemplation\_nn & 0.08533 & 0.131131 \\
donkey\_nn & 0.118224 & 0.097751 \\
edge\_nn & 0.069459 & 0.048838 \\
face\_nn & 0.017471 & 0.017577 \\
fiction\_nn & 0.085575 & 0.065596 \\
gas\_nn & 0.142413 & 0.072828 \\
graft\_nn & 0.126597 & 0.109724 \\
head\_nn & 0.017066 & 0.018624 \\
land\_nn & 0.02214 & 0.027298 \\
lane\_nn & 0.067174 & 0.065796 \\
lass\_nn & 0.106964 & 0.134866 \\
multitude\_nn & 0.065254 & 0.130617 \\
ounce\_nn & 0.122896 & 0.125564 \\
part\_nn & 0.03803 & 0.05156 \\
pin\_vb & 0.109057 & 0.102104 \\
plane\_nn & 0.142745 & 0.074554 \\
player\_nn & 0.146468 & 0.071431 \\
prop\_nn & 0.134356 & 0.14812 \\
quilt\_nn & 0.088169 & 0.090321 \\
rag\_nn & 0.09174 & 0.117761 \\
record\_nn & 0.068515 & 0.047556 \\
relationship\_nn & 0.122012 & 0.063162 \\
risk\_nn & 0.118782 & 0.121072 \\
savage\_nn & 0.059428 & 0.128837 \\
stab\_nn & 0.143965 & 0.123195 \\
stroke\_vb & 0.118052 & 0.089683 \\
thump\_nn & 0.096097 & 0.104708 \\
tip\_vb & 0.092821 & 0.085674 \\
tree\_nn & 0.026129 & 0.035296 \\
twist\_nn & 0.111025 & 0.117444 \\
word\_nn & 0.034585 & 0.046221 \\
\bottomrule
\end{tabular}
\caption{JSD (within each target word and time period) between the frame distributions derived from the lemmatised and raw corpora.}
\label{tab:token_vs_lemma}
\end{table}

\section{Predictions of FTFE model}
\label{sec:FTFE_pred}

\begin{table}[htbp]
\centering
\begin{tabular}{l l l l}
\toprule
\textbf{True Positive} & \textbf{True Negative} & \textbf{False Positive} & \textbf{False Negative} \\
\midrule
prop\_nn  & tree\_nn  & quilt\_nn & ball\_nn \\
graft\_nn & face\_nn  &           & head\_nn \\
stab\_nn  & risk\_nn  &           & ounce\_nn \\
plane\_nn & bag\_nn   &           &          \\
tip\_vb   &           &           &          \\
twist\_nn &           &           &          \\
\bottomrule
\end{tabular}
\caption{True Positive, True Negative, False Positive, False Negative results of FTFE.}
\label{tab:qualitative_grouping}
\end{table}

\end{document}